# Active Data

R. Arthur, V. DiDomizio and L. Hoebel

GE Global Research Center, 1 Research Circle, Niskayuna, NY 12309

{Arthurr, DiDomizio, Hoebel}@research.ge.com

## Abstract

In some complex domains, certain problem-specific decompositions can provide advantages over monolithic designs by enabling comprehension and specification of the design. In this paper we present an intuitive and tractable approach to reasoning over large and complex data sets. Our approach is based on Active Data, i.e., data as atomic objects that actively interact with environments. We describe our intuition about how this bottom-up approach improves designs confronting computational and conceptual complexity. We describe an implementation of the base Active Data concepts within the air traffic flow management domain and discuss performance for this implementation.

## 1. Introduction

Large and complex systems present at least two challenges in trying to apply advanced information systems to the underlying data. First is the computational complexity and related performance impacts of applying polynomial and, even worse, factorial or exponential algorithms to the reasoning process. We use reasoning here in its broadest sense to include such computational sub-disciplines as data mining, computed data updates, cognition and understanding as well as decision-making. The second challenge is that discovery and explanation for humans for decision-making or just simple understanding is a non-trivial task encompassing consideration of user interfaces (presentation), circumscription (domain of explanation and computation), human cognition, relevance and confidence of the reasoning and result. In many complex systems, the act of specification itself may be infeasible in creating the governing logic, exceptions, and design.

Data are widely distributed and come in a myriad of formats. Information changes with time, yet lives on in archives. Humans touch information and mix in opinion, perspective, and nuance of free text. Sources vary in confidence and bias. Storage can be unreliable, prioritized for availability, or require special access. Messages can be repeated, obfuscated, fabricated, and contradictory. Sources for query are created and removed. Data are now being generated at an exponentially increasing rate [Hawking]. In short, data are becoming intractable to digest and store, and more importantly human users require sophisticated tools for querying systems for the information they need.

As an example, we consider flow and congestion detection in the Air Traffic Management (ATM) domain. The complexity includes the number of flights per day (1000's), the number of potential intersections of flight trajectories as well as the number of airports (100's) and the interaction and complexity of the ground-based operations, which includes scheduling of crews and equipment and gate constraints. The domain, in its entirety, includes several NP hard problems. Our examples will focus on just flight routing (the filed flight plan) and identification of congestion in air sectors for flights en-route from origin to destination.

Our approach to this problem is a bottom-up or data-driven approach. We avoid the top-down issues of central control and circumscription of action and explanation as these are both too hard to design and too costly computationally. The small and simple data approach provides an uncomplicated means of circumscription of what data sets are relevant (and considered) for both computation and explanation of results. The issues to consider are (1) whether the hidden complexity transfers to the approach proposed here and (2) what problem types are amenable to the general (monolithic or agent-based) approach as well as our Active Data approach to implementation.

## 2. Architecture, Data and Agents

An agent architecture typically refers to design by functional decomposition (as opposed to a traditional monolithic design). That is, small, specialized, encapsulated programs (agents) provide services to other agents within a medium whereby a community is formed. Within the community, various agent types facilitate exchange such that they form the functional equivalence of the traditional monolithic system.

Agents provide explicit functionality like providing a weather report given a postal code, or provide a transient service like extracting precipitation information from a weather report agent and calculating a visibility code for a map-generation agent. Agents may post requests and capabilities to a global "blackboard" or communicate peer-to-peer. In either case, world-rules govern the basic allocation of resources, inter-agent communication, etc.

Active Data (AD) is an abstraction of an Agent Architecture, whereby the community is composed not of functional elements, but data elements. Notions may be introduced to the community and characterized.

Active Data are then evaluated in the context of a problem space. Within the problem space, data have a context by which to essentially ponder their existence and carry out behaviors such as:

- Aggregating / abstracting based upon peer-data
- Combining with other data
- Alerting other data or known consumers
- Discovery of new information or information gaps
- Self-archival (or deletion)
- Self-replication when "forking" conceptually
- Self-update with newer or higher fidelity information
- Resolution of conflicting information, self-repetition
- Manage exceptions / preferences per world-rules

The base instincts will be to validate truth, improve confidence and detail, and to inhabit an appropriate level of storage. Through such activity, we hypothesize improved storage efficiency, identification of gaps, repetition, and consolidation, leading to potential discoveries through the data themselves.

Additionally, Active Data will self-evaluate in a global context as expressed through constraints and values. Constraints may be based upon resource limitations (e.g. a telescope may be pointed at only one place at a time), temporal limitations (e.g. a package requires 3-5 days for delivery), or similar. Tuple representations and transformation grammars may be applied to guide enforcement of the constraints and evaluation [Carriero].

Active Data differs from a high level definition of agents [Bradshaw], in that agents have a design objective. AD has no objectives, goals or intentions. It does not have beliefs but does have a primitive notion of behavior. AD is immutable at the atomic level with respect to its data value and its (lack of) meaning. On the other hand, characterizing attributes such as Hyperdata (q.v.) of AD can be extremely dynamic.

| Type or *domain* | Attribute (w/computation and controls) | Environment interactions |
|---|---|---|
| Agents | Goals, intentions, beliefs (with methods on self) | Agent behaviors |
| Active Data | Data (value, immutable) | Data behaviors |
| Air Traffic Management | Trajectory (graph nodes, w/constraint propagation) | Sections; calculate times/distances |

Table 1 Active Data, Agents and ATM comparisons

AD and Agents may share certain methods and behaviors, i.e., interfaces and interactions with the environment. The data (value) of AD can come from acts of composition, extraction, abstraction, etc. AD values are not transformable; it may only be deleted, stored or archived after it is removed from active memory. In Table 1 we look at the Agent and AD types and how AD is instantiated in the domain space of Air Traffic Management.

ATM trajectories are updated by constraint propagation (along the graph that represents the air space trajectory). Reasoning is done by the sector over the associated set of trajectory nodes. The ATM atomic Active Data element is the trajectory node associated with a particular sector.

We propose to infuse the data itself with the ability to monitor sources, make new connections, and finally to alert users and environments of new, interesting, relevant information. The activation of data will push many of the tasks of data amalgamation from the user or central-style reasoning system to the data itself. Data in the active system will be responsible for maintaining a constant level of search, propagation and integration. Additionally, data will be capable of some understanding of its relative importance, and will thus move itself to cold storage (if it is no longer needed) or will place alerts (if it is pertinent to a current query) as appropriate.

## 2.1. Architecture

In order to accomplish the challenging goal of imbuing data with such capabilities, it is necessary to first create an environment in which ecology of data might exist. This backplane should be transparent to both the data and the users of the system, much as an operating system is transparent to the users of office applications. We first introduce some concepts.

### 2.1.1. Terms & Definitions

- *Active Data*: Data are the raw materials for all systems professing to act on information, knowledge or intelligence. Often we characterize data as a 'material' that must be gathered, mined, transmitted, analyzed, etc. that is – playing a passive role in the system. Systems, programs, crawlers, agents, and engines are terms for the entities that actively use CPU cycles to make things happen to and from the data. In the case of AD, we instead look to imbue these traditionally passive participants with the basic capabilities to survive in an ecosystem (*behaviors*):
  - *Activity*: ability to gain resources (CPU, memory, bandwidth)
  - *Communication*: ability to interact with other data
  - *Mobility*: ability to change system / storage location
  While we do not literally mean to consider delegation in a strict sense – carried out by functional methods on an object – we pose the question of a conceptual model where this delegation is possible as a means of reducing complexity in managing vast, heterogeneous, multidimensional, multi-ontology data stores.
- *Metadata*: Meta is a prefix that in most information technology usages means "an underlying definition or description." Thus, metadata is a definition or description of data.[1] Typically, metadata is defined within a product or community to describe certain relevant and objective characteristics such as source, size (e.g., Kb or number of elements), or location.
- *Hyperdata*: Unlike metadata, Hyperdata describe dimensional attributes of data beyond the objective. Examples may relate to your perspective of the data source's trustworthiness or bias, or belief to which you feel the data is true, fully detailed, or exposed as common knowledge. Note the changes in Hyperdata over time can be of particular interest.
- *Notion*: To refer to an absolutely generic *thing*, we use the term Notion. It is not a Fact, as it may not be true. It is not an Entity, as it may be an idea or collection. A Notion is quite simply a node in our model, which may correspond to any concept we like – true or false, singular or plural, general or specific. Examples of top-level notions include:
  - *Assumption*: A Notion of posed truth.
    E.g. *Flight 1234 is on time.*
  - *Goal*: A Notion of directional interest.
    E.g. *Least risky investment.*
  - *Hypothesis*: A Notion of explanation.
    E.g. *Rover ate the homework.*
  - *Event*: A Notion of occurrence.
    E.g. *Flight 1234 arrived at 1:10pm.*
- *Ontology*: An ontology is a specification of a conceptualization. That is, an ontology is a description (like a formal specification of a program) of the concepts and relationships that can exist for an agent or a community of agents.[2] Historically, ontologies have been the goal of large standards bodies seeking extensive, detailed models for problem spaces requiring a common taxonomy of vocabulary to allow matching, aggregation and abstraction.

## 2.2. Active Data Characteristics

As we delegate to the data, it should no longer need to be a simple row in a database return, or a simple noun phrase match in an online article with some static metadata attached. The data itself can take on many characteristics as it traverses the system. As a piece of Active Data moves about the system seeking validation or additional relevant information, information is gathered along several axes: truth, confidence, level of detail, public exposure, cache level/storage location, missing information, complementary information, refuting information, and time.
  - *Complementary information* – Information that supports the piece of Active Data. AD is trying to tell a complete story. Any information that it can identify that supports it is useful.
  - *Refuting information* – Information that refutes the piece of Active Data. Necessarily, the data will encounter information that refutes it. This does not mean it immediately discounts the original assumption. Instead, it might factor this refuting information into the total confidence value.
  - *Missing information* – Data that would make the story Active Data are compiling more complete for the user. Given that the AD is piecing together a complete story around this one fact for the user, the AD system can know what kind of information is missing from the complete data set, and could search appropriately.
  - *Truth* – Based on complementary, refuting, and missing information, how true or false the data believes itself to be. Independent of Confidence, Truth ranges from [-1,1] –1 is False, 1 is True. One can have a high confidence in a notion being false.
  - *Confidence* – The degree to which the data is certain of its truth-value. This may be based upon history with the source, corroboration, and subjective measures. E.g. [0,1] 0 is uncertain, 1 is absolute confidence.
  - *Cache level/storage location* – Based on the users' profiles in the system, as well as the truth and confidence values, the Active Data will store itself in the appropriate cache system. Additionally, if the AD is false or not useful, it may destroy itself.
  - *Time* – Each of the data elements have relevance over some period of time. Often the most interesting observations can be gathered from changes in these elements.

Given this more complete view of data, more complete query answers can be generated. Sources users have never explicitly queried can be included in the response, and sources that have low accuracy can be noted as such. Users will have a more complete view of the total data set.

---

[1] http://searchdatabase.techtarget.com/sDefinition/0,,sid13_gci212555,00.html

[2] http://www-ksl.stanford.edu/kst/what-is-an-ontology.html

Some problem characteristics that lend themselves to this paradigm include the data being:

- free of context, or context-introspective
- environment-aware
- locally representative and autonomous
- describable with a simple ontology
- composable into answers
- infeasible to comprehend or specify management in a monolithic paradigm
- opportunistic in forming relationships, aggregations, and value
- pruned from undesirable computations

## 2.3. Operating System for Data

A necessary construct to enable the abstraction of Active Data will be to codify and implement a control structure by which the elements will be managed. Operating systems reside between computer hardware and applications for two main purposes: to simplify implementation of applications and to efficiently use the hardware resources. That is, there are numerous tedious activities common to all applications that can be pushed into the background and effectively taken for granted by merit of running in the environment. Additionally, resources such as CPU cycles, memory, storage, and bandwidth require allocation control to fully exploit their capacity.

| Activity Manager: (Processes / Threads) | Communications Manager: (Signal / Socket) |
|---|---|
| o Create/Load | o Open |
| o Terminate/Abort | o Close |
| o End | o Send |
| o Execute | o Receive |
| o Suspend/Wait | o Block/Wait |
| o Resume | o Poll |
| o Fork | o Confirm |
| o Synchronize | o Semaphore |
| o Interrupt | o Signal |
| Storage Manager: (Memory / Files / Db Rows) | Security Manager: |
| o Create/Insert | o Identification |
| o Open/Allocate | o Authentication |
| o Close/Free | o Authorization |
| o Read/Query | o Credentials |
| o Write/Update | o Permissions |
| o Lock | o Roles |
| o Unlock | o Classification |
| o Commit | o Access List |
| o Rollback | o Virtual Machine |
| o Replicate | o Shared Memory |
| o Distribute | o Protected Memory |

Table 2 Provided background capabilities by the infrastructure

Drawing from the traditional operating system paradigm, and almost without requiring change in the vocabulary, we can consider background capabilities accessed through managers as illustrated in Table 2 above. It is important to distinguish these in that abstractions of their services are taken for granted so that the designer of an Active Data system, like an application developer, has pervasive access to such capabilities.

These design features of an operating system will have direct parallels in the Active Data paradigm, where data:
- Are managed like processes (or threads)
- Manage persistence storage
- Communicate with each other
- Provide mechanisms for data integrity and access control.

The envisioned architecture should draw upon the rich past experience in the field of systems engineering.

In fact, Active Data should lend itself much more powerfully to a system based upon parallel, distributed, or peer-to-peer architectures in that there are highly delegated activities with very small barriers to start-up. Further, such a paradigm shift to handling individual data items as processes on such a fine level of granularity may require developing new methods of computing.

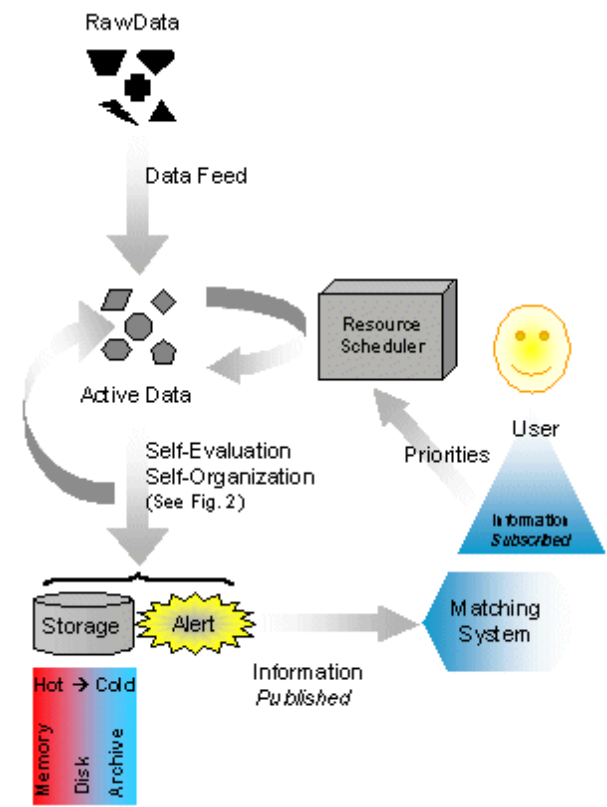

Figure 1: Active Data Lifecycle and Operating System

In Figure 1 we can see a potential architecture for Active Data. Raw Data entering the system would be encapsulated to include requisite metadata and hyperdata. Users would specify points of interest by *subscription* – these could be conceptual, spatio-temporal, etc. Then data would be activated by a Resource Scheduler based upon priorities presented by these points of interest.

Activation would result in tasks such as those seen in Figure 2. For example:

- *Resolution* would result from repeated / redundant data.
- Two or more Active Data could result in a *Hypothesis*, *Aggregated* Notion, etc.

These sorts of self-organizing and self-evaluation activities are driven by the priorities in the Resource Scheduler. In seeking other Active Data with which to compare, activate, etc., we introduce the concept of N-Dimensional Nearness (Figure 3). Consequent to Activation, there will either be a Storage action (for consequent evaluation) or an Alert *published* to the system that matches user requests to provide an output.

Self-evaluation will typically involve update to the hyperdata such as confidence, level of detail, or recency.

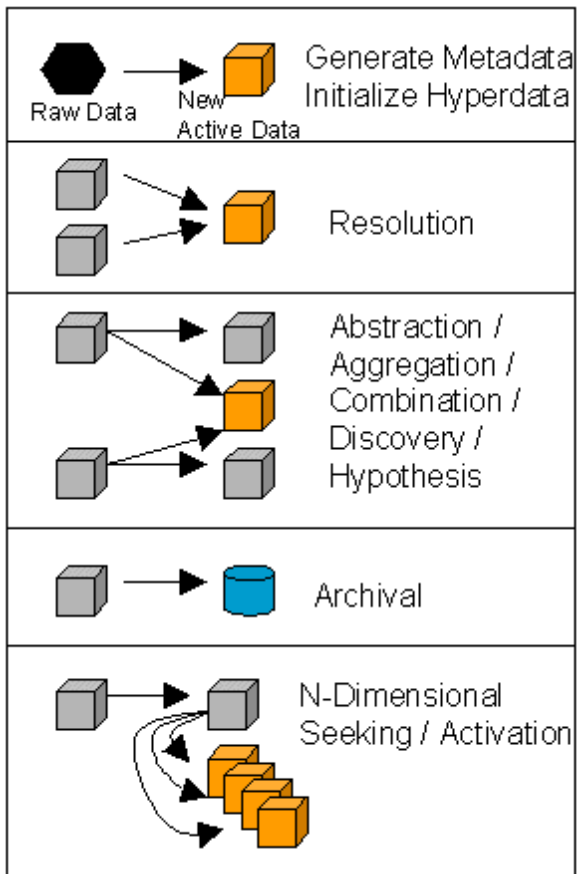

Figure 2: Active Data Activities

An example of Resolution is Figure 2 would be a duplicate news report on election results from CNN and MSNBC. An aggregation may count instances such as "two flights were delayed 16-Oct-2002." An inferred hypothesis might be "The US President is in Paris" - based upon "Air Force One arrived in Paris" and "The US President is on Air Force One." The labels of these activities are not intended to have special or distinct meaning, but rather characterize a group of similar activities.

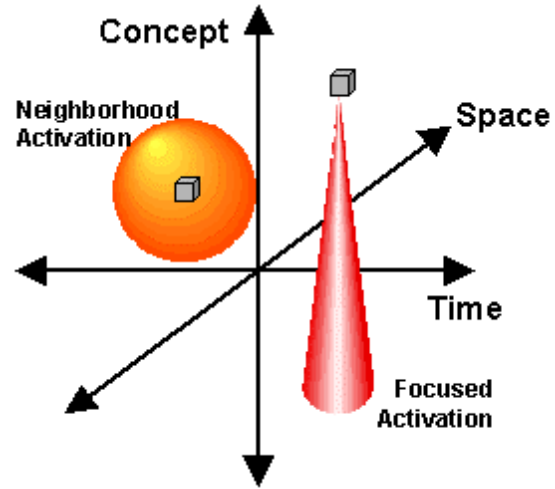

Figure 3: N-Dimensional Nearness - Space, Time, and N Conceptual Dimensions.

Figure 3 depicts a simplification of N-dimensional criteria. As the Resource Scheduler activates AD, these must then consider other AD in the contexts of these dimensions, whether intervals in time, spatial position, or in some sort of ontological categorization. A nearest neighbor search may manifest as a symmetrical range query (or "Neighborhood Activation") or as a directional query across interval constraints in specified dimensions ("Focused Activation").

While Neighborhood Activation will typically be abstract in nature due to multidimensionality, examples of Focused Activation might address one or more intervals such as:
- Temporal: 8-10am, on Tuesdays, before 1990
- Spatial: on I-95, in Bologna, near the Danube
- or Conceptual: Military Operations, Fuel Stations

### 2.4. Multi-participant Model

When the domain must consider data sources such as free text news articles, data may be composed of opinions as well as factual statement. Additionally, systems built to manage data in the context of competitive edge may require deeper understanding of deliberate manipulation by the sources of data.

Truth is often independent of perception, yet perceptions can be important information in decision-making. The field of Game Theory addresses the interaction of decision-makers, based upon behavioral expectations and understanding their strategy. In the real world, we often do not have perfect information among the participants, nor will participants always act rationally, but some fundamentals in competitive behavior still apply, and in the context of data (active or not), the decision-maker with the newest, most accurate, most detailed data has the advantage.

This is particularly true when inferring or aggregating information based upon connections with proprietary information – producing a "window of opportunity" during which one participant possesses a superior position. To evaluate such an opportunity, perception and awareness hyperdata need to be considered in the context of the data set. This means deduce information about what other participants know, whether they know you know or compounded again with confidence and detail over time. Clearly, already vast sums of data now include components that make things even more vast.

Active Data solves problems by making the data itself take on the characteristics of the participants and evaluate relative distance from an equilibrium position. Nash's Non-cooperative Equilibrium refers to a state where all participants are mutually satisfied with their expected result [Nash]. Applying these Game Theoretic approaches, we can impose a mathematical control to activate data to self-evaluate toward a steady state. In the absence of perfect information by all parties, a scoring distance could be formulated which would allow us to assess whether tactical advantage might be achieved, with appropriate notification to the human decision-maker.

We intend to measure the stability of such a system experimentally and address governing functions to manage equilibrium.

## 3. Active Data in Air Traffic Management

As was described above, Active Data has several key characteristics, including mobility, self-awareness, and operating system-like allocation ability. In this section, we investigate how we can currently apply some of these AD characteristics to an Air Traffic Management (ATM) system.

### 3.1. Air Traffic Management – Problem Description

Despite the drop in air traffic since September 11, 2001, it is predicted that by 2020, air traffic will not only return to its previous levels, but will double[3].

The current ATM system will be required to handle new security guidelines as well as this increase in flights. The FAA and the Federal government have concluded that the current ATM will not be capable of handling this volume. The proposed new ATM system will address "…the four most troublesome airspace issues today: the rate of arrivals and departures; en route congestion; local weather; and en route severe weather."[4] In this example, we investigate how Active Data would play into helping alleviate congestion before planes even depart.

The FAA has several goals in mind as it controls air traffic. Some of the factors to consider include: safety, volume, efficiency, and congestion management. These criteria are propagated to the airlines as each airline schedules flights to meet its passengers' requirements. Airlines themselves impose further requirements on their air traffic. For example, they want to deliver the maximum number of passengers safely to their destination along preferred trajectories. This entire set of requirements is further bounded by the space through which trajectories are flown.

Air space in North America is divided into zones, and then further divided into three-dimensional areas called sectors. Planes fly through sectors as they follow their flight plans. Flight plans are requested before take-off and contain waypoints the airplane will fly through en route to its destination. (See Figure 4.)

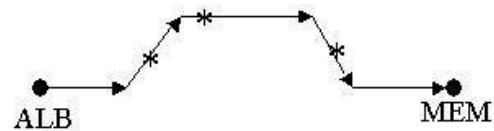

Figure 4: Depiction of a typical flight path. The departure airport is Albany, NY, and the arrival airport is Memphis, TN. The waypoints between the endpoints are specified as asterisks.

Individual sectors have entry and exit points that planes must use as they come into or leave that air space. As can be imagined, each sector has properties such as open or closed air space, wind speed, and current weather conditions. (See Figure 5.)

---

[3] http://www.boeing.com/atm/background/longview.html

[4] http://www.aviationnow.com/content/ncof/ncf_n27.htm

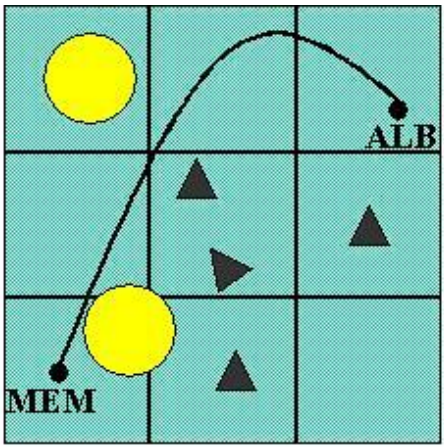

Figure 5: The flight path illustrated in
Figure 4 is laid out over the different sectors. Each green square represents an individual sector. The black triangles represent flights currently en route (which may contribute to congestion in the individual sectors). The yellow circles represent severe weather conditions in the individual sectors and could be represented as neighborhood activation data while the flight path has a focused activation.

Not only can the interior sector properties change over time, but also the sector boundaries themselves can change over time. Additionally, at any given time, planes submitting their flight plans must consider flights currently en route to their destinations. Based on these facts, rules about congestion are determined, numbers of planes are allowed through this air space, and flight plans are approved or adjusted.

The goal of our system is to take the current state of air traffic at a specific time (e.g., planes en route at 06:00 EST), and using historical flight data and current sector definitions, predict where there might be congestion in the next 4 hour time window, thus enabling congestion avoidance.

There are several factors to consider in determining congestion:
- o The physical state of the air space (e.g., physical boundaries of the sector, or the location of closed zones within a sector)
- o The changing condition of weather in an air space
- o Flights currently en route
- o Flights that have requested routes through sectors, and have been added to those sectors as potential flights in the next 4 hour window

When adding a potential flight, there are several things to look at as well:
- o Potential flight paths this airplane could take
- o Weather impacting that flight
- o Any delays that flight will encounter

Given the number of sectors, and the individual conditions each sector manages, a central controller for ATM would have to maintain a vast amount of information and reason globally for each flight. The question is not simply "what flights will be in this sector?" Instead, the question involves understanding the potential delays for every other flight that may be in that sector (due to weather, or take-off delays or previous aircraft delays) as well as anything that may occur in that sector within the next four hours. This involves thousands of flights over all of the sectors, and requires a tremendous amount of reasoning every time the system checks for congestion. For example, consider delays by one flight in its first sector. This adds unusual delays to all of its following sectors, impacting every flight that will be in those sectors, and so on over the length of the delays that extend from there. This is obviously an unwieldy amount of data about which to reason centrally.

If we move to the Active Data architecture, however, most of the propagated effects of this information can be resolved locally. Thus, rather than manage this information from a central controller, the AD model will push the information and decision-making processes to the individual sectors, cutting down on the number of comparisons.

## 4. ATM Architecture

In considering the ATM system and congestion management, we have several types of objects:

*Sectors*
Sectors have boundaries as well as information about the physical space within the boundaries (weather or closed air space). Additionally, sectors contain information about flights currently en route to their destination and flights that have been assigned to fly through that sector within the upcoming four-hour window. Finally, sectors have definitions of congestion for their air space. These definitions may be simply a number of allowable flights, or may involve more complicated algorithms where weather and closed airspaces are included.

*Subsectors*
Sectors are divided into smaller components called subsectors. These portions of sectors have two properties: first they are the smallest unit at which congestion can be defined, and second each sector is completely covered by non-overlapping subsectors.

*Airplanes*
Airplanes have requested and proposed flight information (such as departure time, arrival time, altitude, and air speed) as well as a representation of a proposed trajectory, or flight path.

*Trajectories*
Each flight path is described as a series of waypoints. An airplane requests a specific set of waypoints in its proposed flight plan, beginning with the departure airport, and finishing with the destination airport. Straight-line segments join the waypoints. These segments may be wholly contained in a subsector or may cross two or more subsectors.

The main idea is to understand at what times there will be pieces of trajectories causing congestion in specific subsectors and sectors. In solving this problem, the smallest particle of data with which the system will deal is the amount of time a plane will be in a subsector. Below, we describe how aspects of solving the ATM congestion problem are simplified through using key attributes of Active Data. One can look at this as a non-cooperating zero sum game with each airline (flight) playing a strategy to get preferences (rewards) as it travels a preferred sector route.

*Mobility*

Perhaps the most obvious application of Active Data in ATM is mobility. Given that objects are physically moving throughout the air space, it is easy to picture the data representing these objects as moving through space. However, while the airplane itself moves physically, the segments of the trajectory are moving in the AD model. Here the movement is really the updating of data elements by constraint propagation with the trajectories.

The main question for the ATM system is this:

*Is there congestion if this specific flight with this specific trajectory is added?*

To answer this question, there are several actions the system and the data must take. The trajectory identifies which sectors and subsectors lie along its path as well as what times the aircraft will be within that subsector. It is these time and space intersections with each sector that will be the AD within the ATM system. It is the job of each of these pieces of data to operate appropriately within the context of the Sector (i.e., they must meet the weather constraints, not cause congestion, and not fly through restricted air space).

There are three possible cases when a trajectory segment tries to add itself to a subsector.

**Case 1 – No congestion.** The trajectory path doesn't cause congestion and it adds itself to the subsector. In this case, the Active Data remains in that subsector, waiting for other flights to appear. If, as time passes, there is a case where congestion is caused by another trajectory segment, the set of AD reason together or bargain as described in the case below. Once the flight is en route, the AD moves itself to the en route queue, and finally removes itself from the system when the flight is complete. For example, in the figure below, Sector S has not yet reached its congestion-threshold. (It currently contains four trajectory segments, and given the restricted airspace and weather conditions, it can handle six trajectory segments.) Thus, when flight 3412 adds itself to Sector S, the AD, subsector, and Sector continue normal operations.

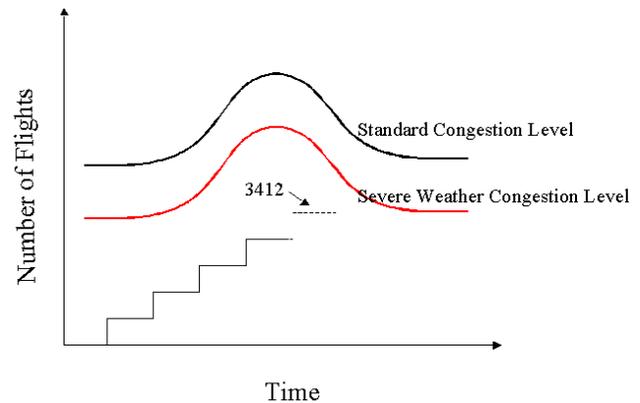

Figure 6 Illustration of a sector where the congestion levels for both calm weather and severe storms have not been reached. The sector is able to add flight 3412 to the set of allowable flights.

**Case 2 – Congestion is reached.** If a segment of the trajectory causes congestion, the AD works with other segments in that Sector context to identify the best solution. (Re-routing the newest addition to the sector may not be the preferred solution.) Once the re-routing plan is determined, the affected trajectory segments remove themselves from their current context and insert themselves into the new subsector. As illustrated in Figure 7 below, Sector S is now congested – it contains 6 trajectory segments belonging to 6 independent airline flight paths. At this point, a trajectory attempts to add itself to Sector S, and the congestion threshold is exceeded. Each of the trajectory segments work together at this point to localize the perturbation to the ATM system. That is, we would prefer to localize the changes to flight plans and trajectories. Each trajectory segment investigates its preferred paths, the least global set of changes that is acceptable to the Sector contexts is identified, and the trajectory segments move themselves to the appropriate subsector environments.

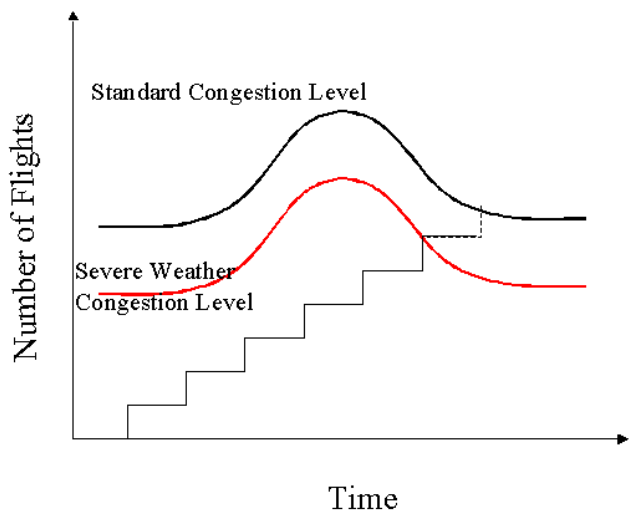

Figure 7 Illustration of a sector where the congestion levels for both calm weather and severe storms have not been reached. The sector is not able to add flight 3412 to the set of allowable flights.

**Case 3 – Context constraint violated.** Congestion being reached within a subsector is merely a specific example of a Sector's constraints being violated. For example, weather or closed or restricted airspace may generate the need for segments to reevaluate their appropriate subsector placement. Again, the Active Data segments within that Sector work to understand the optimal solution.

*Self-reasoning and Introspection*

Another key attribute of Active Data is self-reasoning. As was discussed previously, AD is aware of its surrounding context, and is able to operate within that context. One example of this would be trajectory segments' reacting to weather conditions. Once a segment of the airplane trajectory is placed within a subsector, it must remain aware of the flow of weather in the system.

Weather conditions are represented in each subsector as a set of conditions in that three-dimensional volume. Air speed, storm conditions, and turbulence are examples of what is tracked. Obviously, severe weather storms follow their own trajectories through a series of subsectors. Trajectory segments reason not only about their current subsector's weather, but watch to see the path of a storm, and self-adjust their path as severe weather moves about the system.

## 5. Summary

We have described Active Data, a decentralized concept for a bottom-up approach to reasoning in large complex heterogeneous data environments. We have characterized Active Data and we have shown the basic differences to Agents. We also described an application domain, Air Traffic Management. Finally, we presented our initial implementation work in applying Active Data to that domain for prediction of air space congestion.